# A Cellular Automata based Optimal Edge Detection Technique using Twenty-Five Neighborhood Model

Deepak Ranjan Nayak
Dept. of CSE, College of Engineering and Technology
Bhubaneswar, Odisha
India-751003

Sumit Kumar Sahu
Dept. of CSE, College of Engineering and Technology
Bhubaneswar, Odisha
India-751003

Jahangir Mohammed
P. G. Dept. of Physics, Utkal University
Bhubaneswar, Odisha
India-751004

## ABSTRACT
Cellular Automata (CA) are common and most simple models of parallel computations. Edge detection is one of the crucial task in image processing, especially in processing biological and medical images. CA can be successfully applied in image processing. This paper presents a new method for edge detection of binary images based on two dimensional twenty five neighborhood cellular automata. The method considers only linear rules of CA for extraction of edges under null boundary condition. The performance of this approach is compared with some existing edge detection techniques. This comparison shows that the proposed method to be very promising for edge detection of binary images. All the algorithms and results used in this paper are prepared in MATLAB.

## 1. INTRODUCTION
Cellular Automata (CA), first introduced by Ulam and Von Neumann in the early 1950's with the purpose of obtaining models of biological self-production [1, 2]. Later on, Stephen Wolfram developed the CA theory [3]. The simple structure of CA has attracted researchers from various disciplines. It has been subjected to rigorous mathematical and physical analysis for the last fifty years and its application has been proposed in different branches of science both physical and social.

CA are discrete dynamical systems, and their simplicity coupled with their complex behavior has made them popular for simulating complex systems. Cellular automata offer many advantages over traditional methods of computations:

- All interaction among the cells takes place on a purely local basis that leads to more sophisticated emergent global behavior.

- This simplicity of implementation and complexity of behavior means that CA can be better suited for modeling complex systems than traditional approaches.

- CA are both computationally simple and inherently parallel.

- CA are scalable, as it is easy to upgrade CA by adding additional cells.

- CA continue to perform even when a cell is faulty because the local connectivity property helps to contain the error.

An image can be viewed as a two dimensional CA where each cell represents a pixel in the image and the intensity of the pixel is represented by the state of that cell [5, 6]. The states of the cells are updated synchronously at a discrete time step. So the time complexity to do any image processing task is the least.

Edge detection is a fundamental tool in image processing in the areas of feature detection and feature extraction. In the image, an edge can be defined as the boundary between two dissimilar regions that may result from changes in intensity, color, or texture. The main objective of the edge detection is to identify sharp brightness changes or discontinuities in the brightness level [6]. Cellular automata have been successfully used in the area of image processing for the last couple of years. CA enable fast, parallel computation and have thus found application in image processing as well [9, 10]. There are a number of papers published till date which generally discuss cellular automata for image processing. Also, there were some papers discuss medical image processing based on CA model.

Although there are a number of algorithms have been developed for edge detection but still it is a challenging task to extract proper edges with desirable performance. In this paper we have used two dimensional CA, involves extended Moore neighborhood (twenty five neighborhoods) concept for edge detection. The extended Moore's neighborhood is a 5 ×5 matrix that is used for changing states by comparing differences between a central pixel and its neighbors. Then the linear CA rules are applied to binary images under null boundary condition to get the desired output.

This paper is fall into six parts. In Section 2, the basic concepts of CA and its neighborhood structures are introduced. Section 3 discusses some previous works related to edge detection. The proposed model and algorithm is presented in section 4. The experimental results and comparison is shown in section 5 and 6 respectively. Finally, the conclusion is derived in section 7.

## 2. CELLULAR AUTOMATA
### 2.1 Basic Concepts
Cellular automata are made up from regular grid of cells, where each cell can have finite number of possible states. The state of a cell at a given time step is updated in parallel and determined by the previous states of surrounding neighborhood of cells with the help of a specified transition rule. Thus, the rules of the CA are local and uniform. If all the cells obey the same rule, then the CA is said to be uniform CA. There are one dimensional, two dimensional and three dimensional CA models. One dimensional CA (1D CA) consists of linear arrays of cells whereas in two dimensional CA (2D CA), cells are arranged in a rectangular or hexagonal grid with connections among the neighboring cells, which is depicted in figure 1.

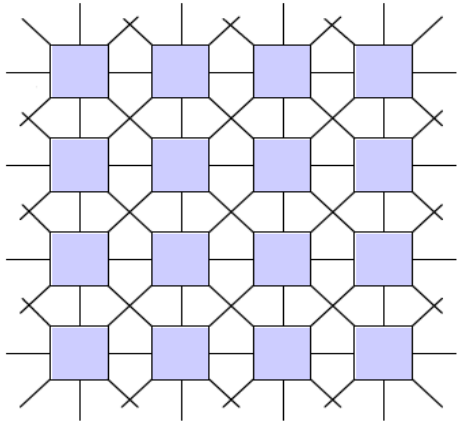

**Fig 1: Structure of 2-D CA**

We can represent a CA with five-tuple, $C = \{L, N, Q, \delta, q0\}$; where $L$ is the regular lattice of cells, $Q$ is the finite set of states, $q0$ is called the initial state and $q0 \in Q$, $N$ is a finite set (of size $n = |N|$) of neighborhood indices such that for all $r \in L$, for all $c \in N: r + c \in L$ and $\delta: Q^n \to Q$ is the transition function.

For a 3- neighborhood 1D CA, the transition function can be represented as

$$qi(t+1) = \delta(qi(t), qi-1(t), qi+1(t))$$

where $qi(t+1)$ and $qi(t)$ denotes the state of the $i^{th}$ cell at time $t+1$ and $t$ respectively, $qi-1(t)$ and $qi+1(t)$ represents the state of the left and right neighbor of the $i^{th}$ cell at time $t$, and $\delta$ *is* the next state function or the transition rule.

As the digital image is a two-dimensional array of m×n pixels, so we are interested in two- dimensional CA model. For a specific problem based on CA, we have to know about the lattice geometry, neighborhood size, boundary conditions, initial conditions, state set and transition rule [9].

The lattice geometry includes the lattice dimension and shape. Here, we have used a square lattice to fulfill our needs as the pixels can be viewed easily in it. Then we have to choose a neighborhood structure through which the cells are updated. The details of neighborhood are described in section 2.2. Most popular boundary conditions are null boundary and periodic boundary conditions which are used when a transition rule is applied to the boundary cells of CA. A CA is said to be a null boundary CA (NBCA) if the extreme cells are connected to logic 0-state and a periodic boundary CA (PBCA) if the extreme cells are adjacent to each other. Initial condition, state set and transition rules are problem dependent, so we will discuss it in section 4.1.

## 2.2 Structure of Neighborhood

The neighborhood of a cell, called the core cell (or central cell), consists of the core cell and those surrounding cells whose states determine the next state of the core cell. There are different neighborhood structures for cellular automata. The two most commonly used neighborhoods are Von Neumann and Moore neighborhood, shown in figure 2 and the parametric position of Moore model is represented in figure 3.

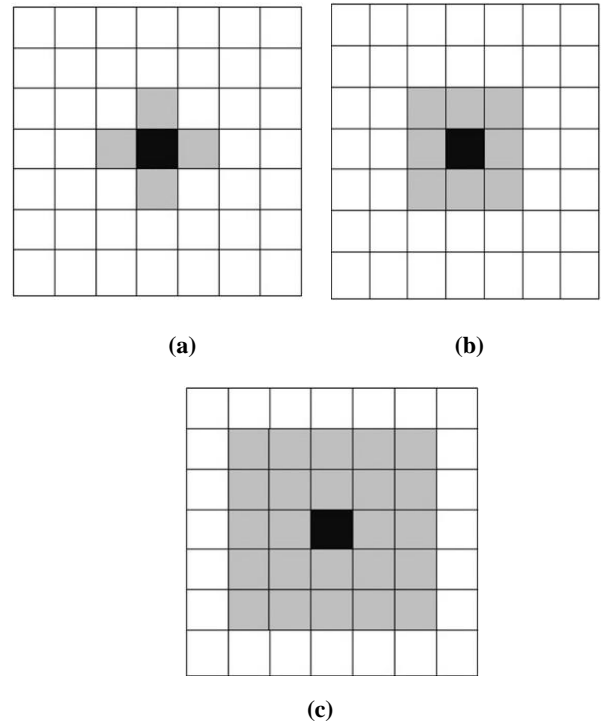

**(a)** **(b)**

**(c)**

**Fig 2: Neighborhood model (a) Von Neumann, (b) Moore, and (c) Extended Moore**

Von Neumann neighborhood has five cells, consisting of the cell and its four immediate non-diagonal neighbors and has a radius of 1. The radius of a neighborhood is defined to be the maximum distance from the core cell, horizontally or vertically, to cells in the neighborhood.

The state of the core cell (i.e. $(i, j)^{th}$ cell) at time $t+1$ depends on the states of itself and the cells in the neighborhood at time $t$. In Von Neumann neighborhood, the transition function is given by:

$$qi,j(t+1) = \delta(qi,j(t), qi,j-1(t), qi,j+1(t), qi-1,j(t), qi+1,j(t)) \quad \ldots (1)$$

Moore neighborhood has nine cells, consisting of the cell and its eight surrounding neighbors and has a radius of 1. Extended Moore neighborhood composed of the same cells as the Moore neighborhood, but the radius of neighbourhood is increased to 2.

The state of the core cell (i.e. $(i, j)^{th}$ cell) at time $t+1$ depends on the states of itself and the cells in the neighborhood at time $t$. The next state function for Moore and extended Moore neighborhood are represented in equation (2) and (3) respectively.

$$q_{i,j}(t+1) = \delta(q_{i,j}(t), q_{i,j+1}(t), q_{i+1,j+1}(t), q_{i+1,j}(t), q_{i+1,j-1}(t), q_{i,j-1}(t), q_{i-1,j-1}(t), q_{i-1,j}(t), q_{i-1,j+1}(t)) \qquad \ldots (2)$$

$$q_{i,j}(t+1) = \delta(q_{i,j}(t), q_{i,j+1}(t), q_{i+1,j+1}(t), q_{i+1,j}(t), q_{i+1,j-1}(t), q_{i,j-1}(t), q_{i-1,j-1}(t), q_{i-1,j}(t), q_{i-1,j+1}(t), q_{i-1,j+2}(t), q_{i,j+2}(t), q_{i+1,j+2}(t), q_{i+2,j+2}(t), q_{i+2,j+1}(t), q_{i+2,j}(t), q_{i+2,j-1}(t), q_{i+2,j-2}(t), q_{i+1,j-2}(t), q_{i,j-2}(t), q_{i-1,j-2}(t), q_{i-2,j-2}(t), q_{i-2,j-1}(t), q_{i-2,j}(t), q_{i-2,j+1}(t), q_{i-2,j+2}(t)) \qquad \ldots (3)$$

| (i-1, j-1) | (i-1, j) | (i-1, j+1) |
|---|---|---|
| (i, j-1) | (i, j) | (i, j+1) |
| (i+1, j-1) | (i+1, j) | (i+1, j+1) |

**Fig 3: Core cell and its neighbor's parametric position in Moore neighborhood model.**

## 3. PREVIOUS RESEARCHES: A SHORT REVIEW

There are several methods of edge detection to deal with different type of edges, each having its own strength. Some methods may work well for one application and may perform poorly in others. Sometimes, experiments are required to find best edge detection techniques for a specific application. Generally, edge detection methods can be grouped into three categories: gradient based edge detection, Laplacian based edge detection, and CA based edge detection techniques. The gradient method detects the edges by looking for the maximum and minimum in the first derivative of the image. The Laplacian method searches for zero crossing in the second derivatives of the image to find edges. The most commonly used gradient and Laplacian based edge detection techniques are Sobel, Robert, Prewitt, LoG (Laplacian of Gaussian) and Canny edge detection operators [22]. Among them, the Canny edge detection operator is known to many as the optimal edge detector. Canny's intention was to enhance the edge detectors that already existed at the time he started his work. The first and most obvious criterion is low error rate. It is important that edges occurring in images should not be missed and that there be no responses to non-edges. The second criterion is that the edge points be well localized that is the distance between the edge pixels as found by the detector and the actual edge is to be at a minimum. A third criterion is to have only one response to a single edge. Based on these criteria, the Canny edge detector first smoothes the image to eliminate noise [4]. In this paper, we have compared our results with these edge detection techniques.

Edge detection based on gradient operators and Laplacian operators requires much computing time. With an increasing demand for high speed real time image processing the need for parallel algorithms instead of serial algorithms is becoming more important. As an intrinsic parallel computational model, cellular automata (CA) can cater this need. Previously, there are different CA models were used for performing edge detection.

Wongthanavasu and Sadananda (2003), proposed a simple CA rule for edge detection [7], and an asynchronous CA model is presented by Scarioni and Moreno in 1998 for the same task [8]. In 2004 Chang et al. introduced a new method of edge detection of gray images using CA [9]. They have considered nine neighborhood structures with periodic boundary condition. An orientation information measure is used to deal with the original grayscale matrix of the image. P L Rosin proposed a different approach on training binary CA for image processing task in the year 2006 [10]. Rather than use an evolutionary approach such as genetic algorithms, a deterministic method was employed, namely sequential floating forward search (SFFS). But the work was only dealt with processing of binary images. Later on, he extends the work to deal with gray images effectively [11]. A new approach for edge recognition based on the combinations of CA and a traditional method of image processing is proposed by Chen and Hao, where they used the concept of boundary operator to represent the state of a cell, and the local rule is defined based on prior knowledge [12]. Lee and Bruce in 2010 propose the concept of using cellular automata and adapted edge detection algorithms for edge detection in hyperspectral images. The authors developed two CAs to analyze the image: an edge detection CA and a post-processing CA (that implements morphological operations for denoising the edges). Results demonstrated the CA method to be very promising for both unsupervised and supervised edge detection in hyperspectral imagery [13].

In the last couple of years, some researchers applied some evolutionary algorithms (such as GA, PSO) to CA for evolving a best rule to perform the edge detection task. Kazar and Slatnia in 2011 used genetic algorithms with CA for image segmentation and noise filtering [14]. In 2012, a meta-heuristic PSO is used by Djemame and Batouche to find out the optimal and appropriate transition rules set of CA for edge detection task. The efficiency of the method was very promising [15]. The concept of fuzzy logic is also somewhat combined with CA for the same task. In 2004, Wang Hong et al. proposed a novel image segmentation arithmetic using fuzzy cellular automata (FCA) [16, 21]. A new improved edge detection algorithm of fuzzy CA is introduced by Ke Zhang et al. in 2007. It has been proved that, the method has great detections effect [17]. More and Patel recently used fuzzy logic based image processing for accurate and noise free edge detection and Cellular Learning Automata(CLA) for enhance the previously detected edges with the help of the repeatable and neighborhood considering nature of CLA [18].

Here we used a novel method of edge detection based on CA, named as TFNCA (Twenty- Five Neighborhood CA). The set of linear rules which we got through several experiments are then applied to TFNCA for detecting edges of an image.

## 4. PROPOSED APPROACH
### 4.1 TFNCA Model
This subsection introduces our proposed TFNCA model for edge detection. For a TFNCA with two state (0 or 1), there are $2^{33554432}$ possible rules exist. Out of them only $2^{25}=33554432$ are linear rules that is, the rules which can be realized by EX-OR operations only and the rest of the $2^{33554432} - 33554432$ rules are non-linear which can be realized by all possible operations of CA. Here, we only consider linear rules for edge detection with null- boundary conditions. The specific rule convention employed here is shown in the figure 4.

| 1048576 | 2097152 | 4194304 | 8388608 | 16777216 |
|---------|---------|---------|---------|----------|
| 524288  | 64      | 128     | 256     | 512      |
| 262144  | 32      | 1       | 2       | 1024     |
| 131072  | 16      | 8       | 4       | 2048     |
| 65536   | 32768   | 16384   | 8192    | 4096     |

**Fig 4: Rule Convention for TFNCA Model**

In this neighborhood structure, the next state of a particular cell is affected by the current state of itself and its surrounding twenty four cells. The central box represents the current cell (in case of an image it is the pixel being considered) and all other boxes represent the twenty four nearest neighbor of that cell. The number within each box represents the rule number associated with that particular neighbor of the current cell. That is if the next state of a cell is dependent only on its present state, it is represented as Rule 1. Similarly, if the next state of a cell is dependent only on its left neighbor, then it is represented as Rule 32 and so on [19, 20]. These twenty five rules are known as fundamental/ basic rules. Using these basic rules all other linear rules are derived which are expressed as the sum of the basic rules. For example, Rule 71, Rule 1097 and Rule 262176 can be expressed as follows:

$$Rule\ 71 = Rule\ 64 \oplus Rule\ 4 \oplus Rule\ 2 \oplus Rule\ 1$$
$$Rule\ 1097 = Rule\ 1024 \oplus Rule\ 64 \oplus Rule\ 8 \oplus Rule\ 1$$
$$Rule\ 262176 = Rule\ 262144 \oplus Rule\ 32$$

Likewise we can express all the possible 33554432 linear rules. Not all the linear rules are applicable to extract the edge of an image. We have found some optimal rules which are giving suitable results and making it comparable with some standard algorithms of edge detection.

## 4.2 Methodology

The methodology encounters the edge detection problem with an extensive use of TFNCA in order to take advantage of its speed and simplicity among other things. Figure 5 illustrates the flow chart of the proposed technique. As every image is considered to be a 2-D lattice of cells, the CA grid width and height is defined by the corresponding image width and height. Here, we consider only the binary images for edge detection; hence each cell can take the values either 0 or 1.

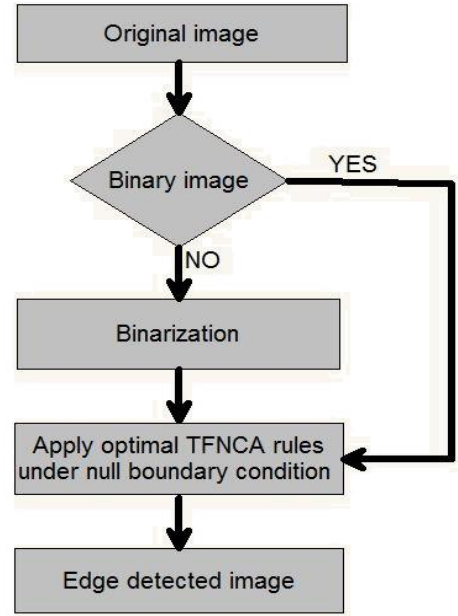

**Fig 5: Flow chart of the proposed methodology**

---
Algorithm: TFNCA ($I_{m \times n}$)

Input: Input image I of size m×n (i.e. this is the initial condition for TFNCA).
Output: Edge detected image.

Step 1: Binarization is needed to convert any image irrespective its size and format to its corresponding binary image with a suitable threshold value. This process is mandatory if the image is not in the binary form. If the input image is in already in binary form then go to next step.

Step 2: Add the null boundary conditions to the image. Then, apply optimal TFNCA linear rules to the image produced at step 1 uniformly. The values of the pixels are updated synchronously with the aim to reduce time complexity for the completion of this task.

Step 3: After the linear rules are successfully applied to the image, an edge detected image is produced.

---

The above algorithm is used in this paper for the desired task. The algorithm mainly follows three steps to get the appropriate results. It is a simple and fast method of edge detection which can be easily implemented in MATLAB or any language.

## 5. EXPERIMENTAL RESULTS

In this section, we have discussed about the results of the proposed algorithm for optimal edge detection. Two grayscale images Lena and Xray of size 256×256 and 302×270 are considered separately to be the input to the algorithm. Then we apply a set of optimal TFNCA rules to the input image and the results are shown in figure 6 and 7. All the results are prepared in MATLAB.

Figure 6 illustrates the edge detection of original grayscale Lena image. Binarization of the original image is presented in figure 6.b. The application of optimal TFNCA rules (Rule 1025, 1040, 1088, 131073, 262145, 262176) and its results are presented in rest of the figures.

Figure 7 shows the result after applying the optimal TFNCA rules to the Xray image. The respective binary image of the input image is demonstrated in figure 7.b by taking a suitable threshold value.

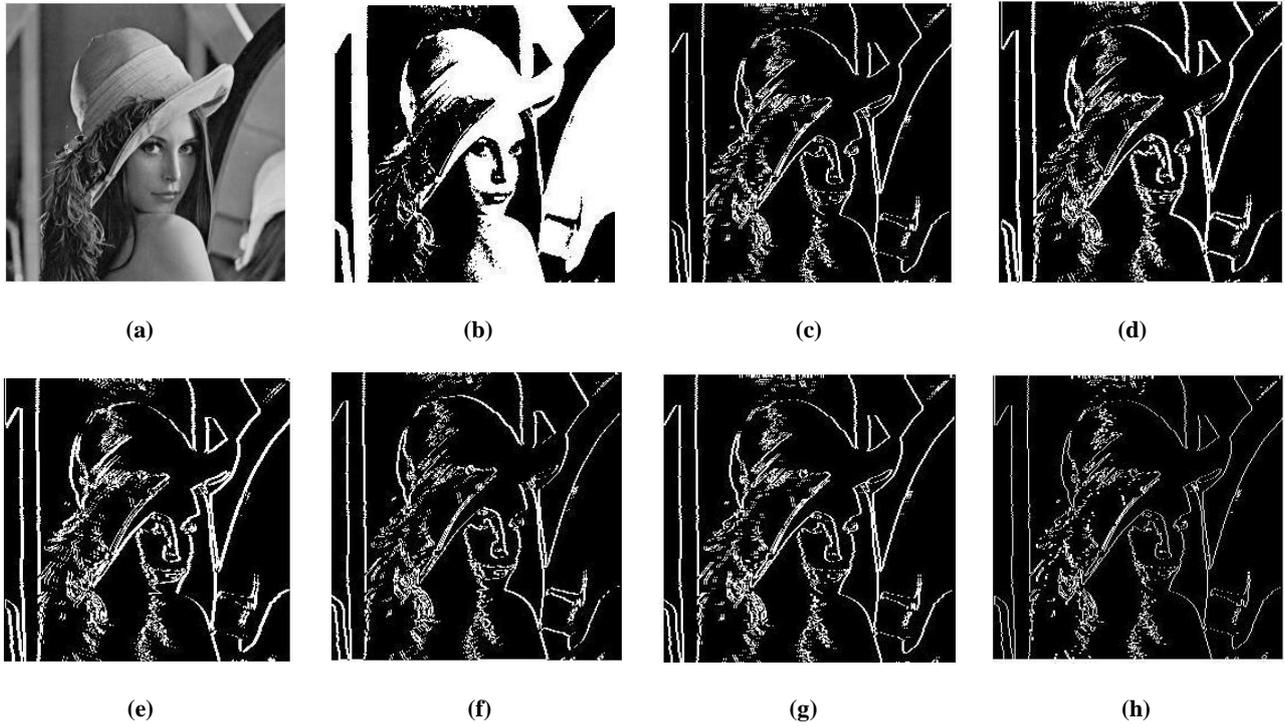

**Fig 6: Edge detection of Lena image of size 256×256 using TFNCA Rule (a) Original Image, (b) Binary Image, (c) Rule 1025, (d) Rule 1040, (e) Rule 1088, (f) Rule 131073, (g) Rule 262145, and (h) Rule 262176**

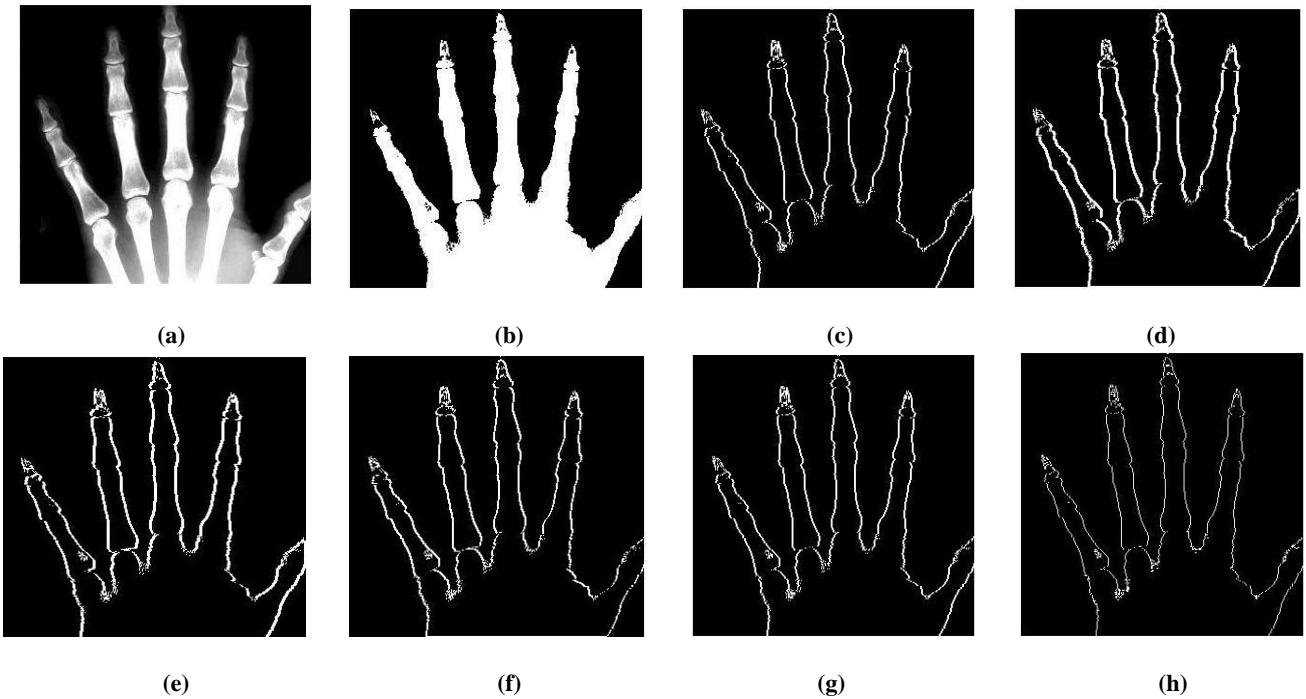

**Fig 7: Edge detection of Xray image of size 302×270 using TFNCA Rule (a) Original Image, (b) Binary Image, (c) Rule 1025, (d) Rule 1040, (e) Rule 1088, (f) Rule 131073, (g) Rule 262145, and (h) Rule 262176**

## 6. COMPARISON

In this section, the experimental results of the proposed method are compared with some traditional edge detection methods namely Sobel, Prewitt, Robert, LoG and Canny method. For this purpose, we consider the same two images Lena and Xray of same size. The results of all the methods are implemented in MATLAB and shown in the figure 8 and 9.

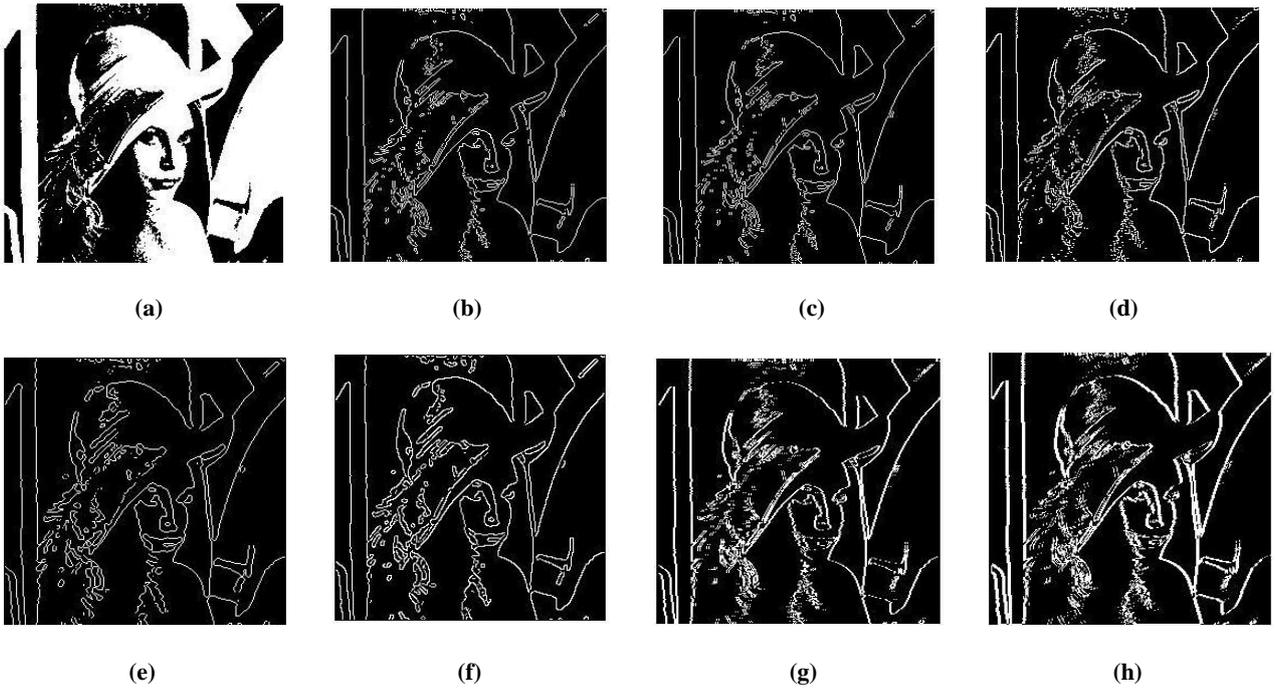

**Fig 8: Edge detection results of Lena image using different methods (a) Original Image, (b) Sobel Method, (c) Prewitt Method, (d) Robert Method, (e) LoG Method, (f) Canny Method, (g) TFNCA (Rule 1025) Method, and (h) TFNCA (Rule 262145) Method**

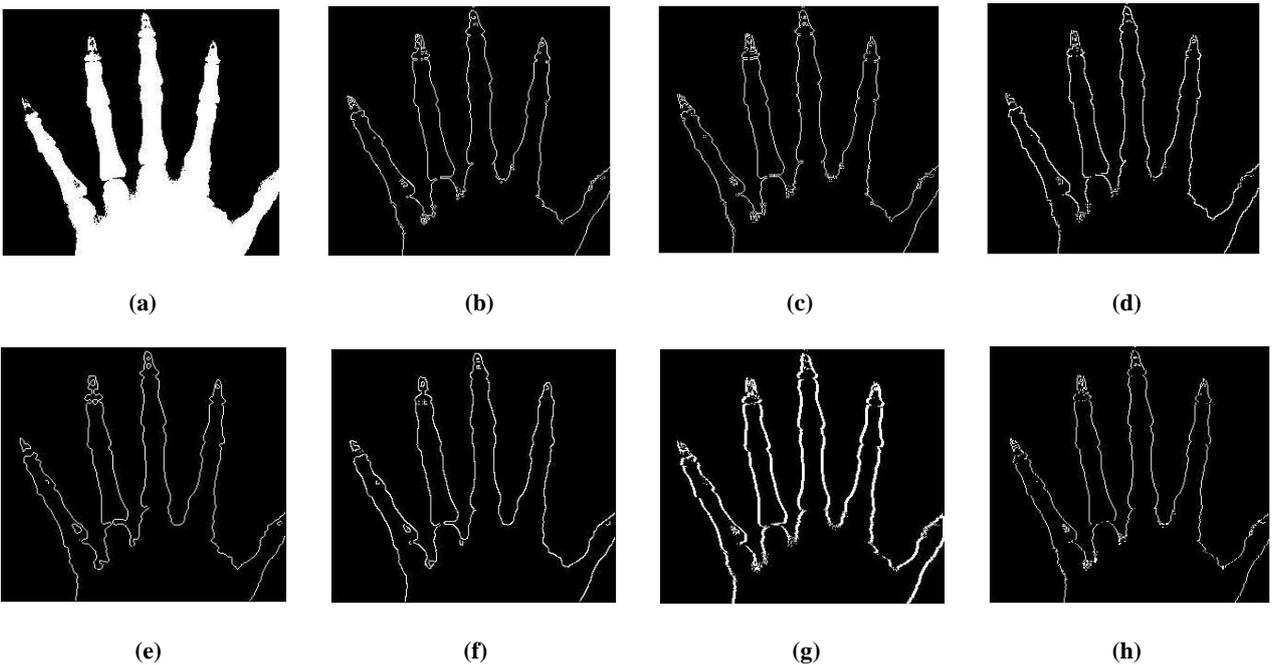

**Fig 9: Edge detection results of Xray image using different methods (a) Original Image, (b) Sobel Method, (c) Prewitt Method, (d) Robert Method, (e) LoG Method, (f) Canny Method, (g) TFNCA (Rule 1088) Method, and (h) TFNCA (Rule 262176) Method**

From the visual images presented in figure 8 and 9, it is clearly shown that the proposed TFNCA rules give optimal results in comparison to others methods. Among all the standard algorithms that we have used for comparison purpose, Canny gives good results. But sometimes Canny produce false edges even though the edge has no existence in the original image. The results demonstrate that the proposed algorithm produces smooth and true edges. In figure 8, we have taken only two rules of TFNCA to compare with others. Due to the paucity of space in this paper we have not added the figures generated by all other optimal rules. Figure 9 presents two other rules, but one can observe from the figure that the result produced by the TFNCA Rule 262176 is nearly same as that of Canny. But all other rules are producing optimal results. Optimal results are defined here in the terms of contrast enhancement that is the results produced by the proposed algorithms have greater contrast than all other algorithms. So the results are more suitable for further analysis.

## 7. CONCLUSION

This paper presents a novel method known as TFNCA for edge detection of binary images. Although the rule space for TFNCA is the large one, still we got some optimal rules for this application through rigorous experiments. The optimal rules produce better results than other existing methods. The proposed algorithm enhances the contrast of the output image and smoothes the edge of the object present in the image. Time complexity of the algorithm is also the least in comparison to others as CA is inherently parallel in nature.

Evolutionary algorithms such as Particle Swarm Optimization (PSO), Genetic Algorithm (GA) and Differential Evolution (DE) can be applied to the proposed method to find the optimal rules so that we can further reduce the computing time. Possible future research direction could be extended to work with gray and color images under different boundary conditions.

## 8. ACKNOWLEDGEMENT

The authors are thankful to Prof. Prashanta Kumar Patra and Dr. Sudhakar Sahoo for their constant encouragement towards this work.